\documentclass[11pt,a4paper]{article}
\usepackage[hyperref]{acl2021}
\usepackage{times}
\usepackage{latexsym}

\usepackage{times}
\usepackage{latexsym}
\usepackage{graphicx}
\usepackage{amsmath,bm}

\usepackage{microtype}

\usepackage{booktabs}
\usepackage{url}
\usepackage{xcolor}
\usepackage{ulem}

\usepackage{float}

\aclfinalcopy 

\title{Exploring Distantly-Labeled Rationales in Neural Network Models}

\author{
    Quzhe Huang, 
    Shengqi Zhu, 
    Yansong Feng\thanks{\;\;Corresponding author.}~~,
    Dongyan Zhao \\
    Wangxuan Institute of Computer Technology, Peking University, China\\
    The MOE Key Laboratory of Computational Linguistics, Peking University, China\\
    {\tt \{huangquzhe, zhusq, fengyansong, zhaody\}@pku.edu.cn} \\
}

\date{}

\begin{document}
\maketitle
\begin{abstract}
Recent studies strive to incorporate various human rationales into neural networks to improve model performance, but few pay attention to the quality of the rationales.
Most existing methods distribute their models' focus to distantly-labeled rationale words entirely and equally, while ignoring the potential important non-rationale words and not distinguishing the importance of different rationale words. In this paper, we propose two novel auxiliary loss functions to make better use of distantly-labeled rationales, which encourage models to maintain their focus on important words beyond labeled rationales (PINs) and alleviate redundant training on non-helpful rationales (NoIRs). Experiments on two representative classification tasks show that our proposed methods can push a classification model to effectively learn crucial clues from non-perfect rationales while maintaining the ability to spread its focus to other unlabeled important words, thus significantly outperform existing methods.

\end{abstract}

\section{Introduction}
\label{sec:intro}

Recent studies have shown an increasing interest in incorporating human knowledge into neural network models \cite{xu2018semantic,vashishth2018reside,luo2018marrying,li2019augmenting,re2020emnlp}. For many natural language processing (NLP) tasks, such domain knowledge often refers to salient words annotated by human experts, which are also called \textit{rationales}. Table~\ref{fig:intro} (top) shows an example of  expert-annotated rationales for sentiment analysis, which highlight noteworthy tokens and score the contributions of these tokens.
The detailed annotations reflect the importance of these words from the expert annotator's viewpoint and are expected to help training better sentiment classification models.

\begin{table}[]
  \setlength{\tabcolsep}{1mm}
  \centering
  \small
  \begin{tabular}{p{0.14\columnwidth}p{0.80\columnwidth}}

  Expert Labeled & \underline{Painful$_{[0.1]}$} to watch,
  \underline{but$_{[0.7]}$} viewers willing to take a chance will be \underline{rewarded$_{[0.6]}$} with two of the year's most  \underline{accomplished$_{[0.6]}$} and  \underline{riveting$_{[0.9]}$} film performance.  \\ \\
  Distantly\ \ \  Labeled & \uwave{Painful} to watch, but viewers willing to take a chance will be rewarded with two of the year's most accomplished and \uwave{riveting} film performance.

  \end{tabular}
  \caption{An example of rationale annotation for sentiment analysis. \underline{Words} in underline are rationales annotated by human experts, and \uwave{words} in wavy underline  are annotated via sentiment lexicon matching. 
  Numbers in [] are salience scores labeled by experts.
  } 
  \label{fig:intro}
 
  \end{table}

Nonetheless, careful, case-by-case rationale annotations inevitably involve large amounts of manual efforts, and are often extravagant or not even available. In practice, distantly-labeled rationales serve as a plausible alternative. Instead of labelling case by case, annotators could design heuristic rules to generate rationales for the whole dataset. For instance, in sentiment analysis, annotators can collect words with strong sentiment polarity (positive or negative) to construct a sentiment lexicon,  with which they can  automatically annotate rationales in a short time through word matching, such as \textit{Painful} and \textit{riveting} in the bottom case of Table~\ref{fig:intro}. 
When comparing the bottom annotation with the top one, we should admit that  the automatic annotations are not perfect, where they indeed include useful clue words towards sentiment prediction. But there should be differences of importance  among those  automatically annotated words, e.g., compared to \textit{Painful}, \textit{riveting} is more important to decide the sentiment of the sentence, and several important clues are still missing, e.g., \textit{but}, \textit{accomplished}, \textit{rewarded}, etc.
Distantly-labeled rationales drastically reduce the cost of generating precise case-specific annotations while preserving a certain degree of reliability, thus are widely used.

However, just as researchers apply auxiliary measures to enforce higher concentration on distantly-labeled rationales and expect substantial model gains, potential flaws in the quality of those rationales quietly arise to hinder the model from benefiting from human priors. Specifically, as discussed in the sample annotations, we find there are, among others,  mainly two types of quality issues lying in distantly-labeled rationales:

\textbf{Insufficiency}. Since distantly-labeled rationales do not include case-specific checking and only contain universally helpful words according to predefined rules/lexicons, such rationales may not provide sufficient supporting evidence in individual cases, and more information from non-rationale words may be necessary towards the final classification.
Given an instance with distantly-labeled rationales, we call the unlabeled words that are contributing to the final prediction as  \textit{Potential Important Non-rationales}, or PINs for short, e.g., \textit{but} and \textit{rewarded} in the bottom of Table~\ref{fig:intro}.

\textbf{Indiscrimination}.  Although distantly-labeled rationale words are often universally helpful, given a specific context, different rationale words may exhibit varied importance. If those distantly-labeled rationales are applied in a 0-1 form to all instances and treated equally important, the tremendous diversity of actual importance in individual cases is just ignored. We refer to the distantly-labeled rationale words that are not helpful in a specific instance as \textit{Non-Important Rationales}, or NoIRs for short. 

Although many existing works have attempted to incorporate automatically-obtained rationales in different ways and achieved promising results in various applications \cite{liu2017exploiting,nguyen2018killed,ghaeini2019saliency,liu2019incorporating}, 
they do not explicitly examine the quality issues of distantly-labeled rationales,  nor formally consider them during modeling, except \cite{featurefeedback2017}, which incorporate vague  feature feedback into a linear classifier. 
On the one hand, most existing  methods try to apply strict constraints to require model focus to conform to rationales, often encouraging those words to share all the model focus \cite{nguyen2018killed,liu2019incorporating}. However, as distantly-labeled rationales are often insufficient to draw correct conclusions, the rigid requirements may turn out to incorrectly ignore the PINs. On the other hand, rationale words are often expected to share  equal importance, which is not the case in practice and can falsely lift the focus on NoIRs.

In this paper, we seek better ways to exploit distantly-labeled rationales, and analyze to what extent the aforementioned quality issues can be alleviated with our methods. We propose two novel gradient-based schemes, namely Order Loss and Gate Loss, to handle the insufficiency and indiscrimination problems, respectively. Order Loss presents a relaxed constraint on rationales by requiring them to have higher gradients than non-rationales, instead of occupying the entire model focus. Gate Loss introduces an early stop mechanism, which prevents over training that enhances the significance of non-helpful rationales. 
We evaluated our methods on two NLP tasks, sentiment analysis and event detection, and  the experimental results show that our methods can better exploit non-perfect distantly-labeled rationales, paying attention to PINs while avoiding over-training on NoIRs, thus outperforms competitive counterparts. 

Our main contributions are as follows:
    (1) We formally address the quality issues of distantly-labeled rationales, namely insufficiency and indiscrimination, and propose two novel loss functions to push the model training process while taking the potential important non-rationales (PINs) and non-important rationales (NoIRs) into account.  The two new losses can also be jointly used and lead to further improvement. 
    (2)  We conduct comprehensive evaluations on two classification tasks and our analysis shows that  our proposed methods can better deal with automatically-annotated rationales, even in a lower quality. 

\section{Word Salience}

Before elaborating on our proposed methods, we first introduce the definition of \textit{word salience}, a measure of the importance of words, which is widely applied in previous works \cite{luo2018marrying,nguyen2018killed,jin2019towards}

Given a model $f$ and an input word sequence $\mathbf{x}=(x_1, x_2, ..., x_n)$, the word salience is a vector $\mathbf{s} = (s_1, s_2, ..., s_n)$ that denotes the importance of every word in $\mathbf{x}$, where $s_i$ indicates how much $x_i$ contributes to the model $f$.

Prior works have explored different methods to determine word salience \cite{ ribeiro2016should,jin2019towards}. Among them, we choose gradient-based methods since they are model-agnostic and easy to obtain. Moreover, since gradient-based word salience is differentiable with respect to model parameters, taking it as part of the objective makes it more convenient to optimize the loss.

For a function $f$, the magnitude (absolute value) of its gradients with respect to input $\mathbf{x}$ indicates how sensitive the final decision is to the change of $\mathbf{x}$ \cite{li2016visualizing}.
In most NLP settings, the gradient of a word is the sum of gradients for each dimension of word embeddings. Formally, the gradient of an input word $x_i$ to a function $f$ can be calculated as:

\vspace{-1em}
\begin{equation}
    g_i = \left\| \frac{\partial f}{\partial x_i} \right\|_1
    \label{eq:gi}
\end{equation}
where $\|\cdot\|_1$ is the $L_1$ norm that sums up the absolute value of gradients over the embedding dimensions.

For gradient-based methods, we use the normalized gradients to calculate word salience, which represents the proportion of a word's contribution in a sentence:

\vspace{-1em}
\begin{equation}
    s_i = \frac{g_i}{\sum_{j=1}^n g_j}
    \label{eq:norm_gi}
\end{equation}

There exist more complicated gradient-based methods for calculating word salience \cite{sundararajan2017axiomatic}. Here, we base the salience on the vanilla gradient method, for the following reasons: 1) It is simple yet sufficiently effective to represent word salience \cite{ross2017right}; 2) The calculation cost of the vanilla version is minimal among all gradient-based methods.

\section{Our Methods}

To incorporate human rationales into neural models, most existing works introduce an auxiliary loss to impel the neural network model to put more emphasis on rationale annotations. Formally, for a multi-class classification problem, the joint objective can be formalized as:
\begin{equation}
    L_{joint} = L_c(\mathbf{x}, y) + \lambda L_a(\mathbf{s}, \mathbf{z})
\label{full_loss}
\end{equation}
where $L_c$ is the classification loss based on input sentence $\mathbf{x}$ and ground truth label y, and $L_a$ is a constraint function which conforms word salience $\mathbf{s}$ with a binary vector of rationale labels  $\mathbf{z}=(z_1, z_2, ..., z_n)$ where $z_i$ is 1 if $x_i$ is important, otherwise, $z_i$ is set to 0. $\lambda$ is the hyper-parameter controlling the weight of the auxiliary loss.

We start with a discussion on a Base Loss currently in use, which suffers from the insufficiency and indiscrimination of non-perfect rationales. Alternatively, we introduce two methods, namely Order Loss and Gate Loss, which help models to minimize the influence of NoIRs and leave enough space for PINs as well.

\subsection{Base Loss}
Most previous works consider all rationale words as carefully annotated and flawless, without taking the quality issues into account \cite{liu2017exploiting,liu2019incorporating}. 

Generally, their main assumption could be written as: 
   \textbf{A1: } \textit{All rationales contribute equally to the model, while other words should not contribute.}
According to this assumption, the salience of every rationale word should be equal to each other, which is $\frac{1}{k}$ for a sentence with $k$ annotated rationale words. Meanwhile, the salience of non-rationale words is set as 0. When using $L_2$ norm to measure the difference between the current word salience and the expected values (0 or 1) for each word, we can write the constraint loss as: 

\vspace{-0.2em}
\begin{equation}
    \begin{aligned}
        L_{a\_base} &= \sum_{z_i = 1} \left(s_i - \frac{z_i}{\sum_{j=1}^n z_j}\right)^2 \\
        &= \sum_{z_i = 1} \left(s_i - \frac{1}{k}\right)^2
    \end{aligned}
    \label{eq:la_haha} 
\end{equation}
where $s_i$ is the salience of rationale word $x_i$ and $\frac{z_i}{\sum_{j=1}^n z_j}$ is the expected value for $x_i$, which equals $1/k$ for a sentence with $k$ annotated rationale words, since $\mathbf{z}$ is a binary vector.

Although this loss exhibits a feasible way to allow rationales to receive higher concentration, it also has two distinct shortcomings.  Firstly, rationales often possess varied importance in real-world cases, which makes it improper to strictly require equal concentration.
Second, for the important words not covered in rationales, they are totally ignored and cannot contribute to the prediction.

\subsection{Order Loss: Exploiting PINs} 

For distantly-labeled rationales, 
\textbf{A1} pushes the classification model not to make use of potential important words outside rationale annotations, and squeezes them to  receive little focus.
To make better use of these PINs, we seek to relax the restrictions between rationales and non-rationales, and propose the following assumption as an alternative:
\textbf{A2:}  \textit{Rationale words should get more focus than non-rationales.}
    \label{assump:order}
Based on this assumption, we can directly build up a formal restriction between any pair of rationale word and non-rationale word: 
\vspace{-.5em}
\begin{equation}
    s_i > s_j \quad \forall x_i \in S_R, \forall x_j \in S_N
    \label{res_full}
\end{equation}
where $S_R$ is the set of annotated rationale words, $S_N$ is the non-rationale set, and $s_i$ and $s_j$ are the salience of words $x_i$ and $x_j$, respectively. This restriction enumerates all the possible pairs of annotated rationale and non-rationale words, and involves massive computation. 
For a sentence of length $n$ with $k$ labeled rationale words, constraining the above order relationship (Eq.~\ref{res_full}) leads to considering $k(n-k)$ terms in the auxiliary loss. This is expensive for longer sentences with sparse rationale annotations. It is worth looking for a more efficient constraint method that is irrelevant to sentence length and only involves rationale numbers.

However, if we know the maximum value $\max s_j$ in $S_N$ in advance, most of the comparisons in (Eq.~\ref{res_full}) can be omitted, because requiring an $s_i$ to be greater than every $s_j$ is equivalent to requiring $s_i > \max s_j$. Therefore, we can simplify the restriction in Eq.~\ref{res_full} to: 
\vspace{-0.7em}
\begin{equation}
    s_i > \max s_j \quad \forall x_i \in S_R, \forall x_j \in S_N
\end{equation}

Since salience can vary enormously in orders of magnitude, it is hard to determine $\lambda$ in Eq.~\ref{full_loss} and converge to a stable state if we just calculate the loss regarding the difference between $s_i$ and $\max s_j$. 
In order to obtain a loss that is insensitive to the magnitude of salience, we adjust the restriction to an equivalent form:

\vspace{-0.5em}
\begin{equation}
    \frac{s_i}{\max s_j} > 1  \quad \forall x_i \in S_R, \forall x_j \in S_N
\end{equation}

And its corresponding auxiliary loss can be written as:
\vspace{-0.7em}
\begin{equation}
    L_{a\_{order}} = \sum_{z_i = 1} \left( min\left(\frac{s_i}{ \underset{z_j = 0}{max}~s_j} - 1, 0\right) \right)^ 2
    \label{eq:la_new}
\end{equation}
where $\underset{z_j = 0}{max}~ s_j$ is the maximum salience among all non-rationale words. 
The $min$ function guarantees that no restrictions will be applied as long as the maximum salience of non-rationale words is smaller than any rationale word.

\subsection{Gate Loss: Handling NoIRs}
Distantly-labeled rationale words may vary dramatically in quality. Non-helpful rationale words may incorrectly attract the model focus, which may confuse the model and affect its performance.
To address this problem, we thus make a new assumption, which prevents the model from overly focusing on the rationales that are not helpful:
    \textbf{A3: } \textit{Only part of the rationales, or crucial rationales, should attain higher focus.} This could encourage a model to give little focus to certain annotated rationale words that are identified as non-helpful during training. 

Since Base Loss explicitly requires an equal focus on all rationale words, the model will drag the salience of rationales to be equal after long periods of training. 
This is not expected for distantly-labeled rationales, as some of them may not be helpful.
We expect an adaptive early-stop mechanism for such losses 
in order to prevent over-training on those non-helpful rationales.

Specifically, we consider halting the auxiliary constraining process when rationale words in an instance have gained adequate focus in total. This indicates that some rationale words are already identified as important during training. 
As those rational words are sufficient towards final classification, there is no need to enhance the others.
In contrast, for instances where the total focus for all rationale words remains at a lower level, they should possess a higher priority in the remaining training process.

To this end, we add a \textit{Gate} term to Base Loss to form \textit{Gate Loss}, in order to adaptively determine whether to skip the gradient constraints for the current instance:
\begin{equation}
    L_{a\_gate}=Bern(1 - \sum_{x_i \in S_R} s_i)\sum_{x_i \in S_R} (s_i-1)^2
    \label{eq:gate}
\end{equation}
where \textit{Bern(p)} is the Bernoulli distribution with parameter $p$.\footnote{We have also tried other common methods besides Bernoulli distribution, and the results are shown in the Appendix.}
The Gate term can be similarly attached to Order Loss to jointly apply the two methods:
\begin{equation}
    L_{a\_gate+order}=Bern(1 - \sum_{x_i \in S_R} s_i) L_{a\_order}
    \label{eq:alll}
\end{equation}

With this term Eq.~\ref{eq:gate}, constraints are given less and less opportunity as the sum of rationale salience rises. The more focus the current rationales receive in total, the less likely the instance will be further trained on. Thus, the Gate term acts as a gate for sentences with both helpful and non-helpful rationales: as the most helpful rationale words quickly stand out and take up a higher proportion in salience, rationales in these sentences will have lower chances to receive training in the future iterations. In other words, the Gate term allows the model to focus on instances whose rationale words are not well modeled.

\section{Experiments}
We evaluate our methods on two sentence classification tasks, sentiment analysis and event trigger detection, on  Stanford Sentiment Treebank (SST) and ACE-2005, respectively, which have been considered as a suitable testbed to investigate how additional rationales can help to improve a base model. 

\textbf{Stanford Sentiment Treebank (SST)}~\cite{socher2013recursive} 
includes 10,662 sentences tagged with sentiment on a scale of 1 (most negative) to 5 (most positive). We filter out neutral instances and divide
the remaining sentences into positive (4, 5) and negative (1, 2), making it a binary classification task. There are 6920 sentences in training set, 872 sentences in validation set and 1821 sentences in test set.
In SST, words are labeled with 5 levels of sentiment polarity. We take the words with extreme positive polarity (label 1) or negative polarity (label 5) as our sentiment lexicon, which is used to automatically annotate rationale words in each sentence. 56.7\% training instances have at least one rationale word. There are 0.85 annotated rationale words per sentence on average, and the average sentence length for training is 19.3 words.

\textbf{ACE-2005} \cite{ace2005} is an Event Detection (ED) Dataset. Following previous works in event detection{~\cite{nguyen2015event}},  we consider event trigger detection as a classification task. That is, for every token in a given sentence, we aim to predict whether the current token is an event trigger or not. Here, we do not consider identifying event types and formulate it as a binary classification task for ease of exposure.  

Previous studies show that trigger words are strong, universal features that can indicate events of specific types. Therefore, in each sentence, we automatically label a word as rationale if and only if it has been labelled at least once as a trigger in the training set.  
We use the same split as \cite{DBLP:conf/acl/JiG08}, with 14,849 sentences for training, 836 for validation, and 672 for testing. 
88.7\% of training sentences have been annotated with at least one rationale word. On average, there are 4.66 rationale words per sentence.

\paragraph{Evaluation Metrics}
Following previous works, we use accuracy (\textbf{Acc}) and F1-scores (\textbf{F1})  as the evaluation metrics on SST. 
We use F1-score as the only metrics on ACE-2005 and do not examine accuracy, since this dataset is extremely unbalanced, where a model predicting all instances into negative can achieve over 97.5\% Acc. We run each setting 5 times and report mean and standard deviations.

\paragraph{Implementation Details}
Our basic classification model is a convolutional neural network (CNN)\cite{ghaeini2019saliency}. The input tokens are first transformed to word embeddings, which are 300-dimension Glove vectors \cite{DBLP:conf/emnlp/PenningtonSM14} in SST, and the combination of a 300-dimension Glove embedding and a 50-dimension entity (originally labeled) embedding in ACE-2005. Then, a convolution layer with 200 kernels, viz. 50 kernels with width 2, 3, 4 and 5 respectively, is used to extract local features, followed by a feed forward neural network to gain hidden representations of words. We then calculate the sentence representation using the attention mechanism, and feed it into a softmax regression to obtain estimated probability distribution.
All activation functions are $tanh$, the dropout rate is 0.5, and the batch size is 512 for SST and 256 for ACE. We optimize the model with Adam \cite{kingma2014adam} with learning rate = $10^{-3}$, $\beta_1=0.9$, $\beta_2=0.999$ and $\epsilon = 10^{-8}$. The $L_2$-normalization rate is set to $10^{-4}$.

For instances without any annotated rationale words, we do not apply auxiliary losses to them.

\paragraph{Comparison Methods}
Besides the base CNN model, we compare our methods with 2 recent works that combine the same CNN architecture with additional rationales:
\textbf{CNN:} the vanilla CNN classifier trained with the cross-entropy loss.
Saliency Learning (\textbf{SL}): \citet{ghaeini2019saliency}  proposes a broad constraint that requires all rationale words to have positive gradients. 
Integrated Gradient Attribution (\textbf{IGA}): \citet{liu2019incorporating}  use the Integrated Gradient \cite{sundararajan2017axiomatic} to calculate the attributions of a classification model, and force the model to focus on rationales by restricting their attributions to be 1 , where the word attribution is similar to word salience in our work.

\begin{table*}[]
\centering
\small
\setlength{\tabcolsep}{12pt}
\begin{tabular}{@{}lcccccc@{}}
\toprule
               & \multicolumn{2}{c}{Accuracy (SST)}                & \multicolumn{2}{c}{F1-score (SST)} & \multicolumn{2}{c}{F1-score (ACE-2005)} \\ \midrule
Model          & Mean + Std. & \multicolumn{1}{l}{Sig. p} & Mean + Std.   & Sig. p   & Mean + Std.          & Sig. p   \\ \midrule \midrule
Baseline       & 0.847  $\pm$ 0.002 & -                      & 0.851 $\pm$ 0.003   & -                   & 0.698 $\pm$ 0.004   & -            \\ \midrule
SL       & 0.849  $\pm$ 0.003 & -                      & 0.851 $\pm$ 0.004   & -                   & 0.704 $\pm$ 0.004   & -            \\
IGA & 0.848  $\pm$ 0.002 & -                      & 0.852 $\pm$ 0.002   & -                   & 0.703 $\pm$ 0.003   & -            \\ \midrule
+ Base Loss    & 0.851  $\pm$ 0.004 & -                      & 0.854 $\pm$ 0.004   & -                   & 0.705 $\pm$ 0.005   & -            \\
+ Gate Loss    & 0.852  $\pm$ 0.003 & -                      & 0.854 $\pm$ 0.005   & -                   & 0.714 $\pm$ 0.006   & 0.044        \\
+ Order Loss   & \textbf{0.862}  $\pm$ 0.003 & 0.008         & \textbf{0.862} $\pm$ 0.003   & 0.041        		 & 0.715 $\pm$ 0.005   & 0.032        \\
+ Gate + Order & 0.861  $\pm$ 0.004 & 0.013                  & \textbf{0.862} $\pm$ 0.003   & 0.047        		 & \textbf{0.726} $\pm$ 0.008  & 0.002 \\
\bottomrule
\end{tabular}
\caption{Performance of our approaches on two dataset with CNN as base model. Saliency Learning and IG Attribution are our implementations of two previous gradient constraint methods. \textit{+Base}, \textit{+Order}, \textit{+Gate} stand for models with corresponding auxiliary losses, and \textit{+Gate+Order} is Order Loss combined with the Gate term. \textit{Sig. p} columns report the p-value of t-test with \textit{+Base Loss}.}
\label{tab:main_result}
\end{table*}

\subsection{Main Results}
Table~\ref{tab:main_result} shows the performance of different methods on SST and ACE-2005. 

We first notice that  previous approaches, both Saliency Learning and IG Attribution, perform slightly better than the baseline CNN classifier, without significant improvement. This is not surprising, since in our setup, the rationale annotations are automatically collected, far from perfect compared to expert-annotated ones. Although both SL and IGA push the classification model to focus on those rationales, neither of them takes into account the quality issues of distantly-labeled rationales, i.e., insufficiency and indiscrimination, thus it is difficult for them to bring more significant improvement regarding vanilla CNN. The Base Loss method also poses strong emphases on the rationale words without considering PINs. It can bring a bit more improvement than SL and IGA, though not significant enough.  When we push the classification model to consider the different importance of these non-perfect rationale words, our Gate Loss method obtains more significant improvement on ACE-2005. When formally considering to spread the model focus to PINs, our Order Loss method obtains significant improvement, 1.1\% and 1.7\% improvement in F1 than vanilla CNN on SST and ACE-2005, respectively. 

Now we look closer at the performance of our proposed methods. On ACE-2005, applying Order Loss and Gate Loss  can both significantly outperform vanilla CNN in F1-scores, by $1.7\%$ and $1.6\%$, respectively. This is more than twice the improvement gained by the Base Loss ($0.7\%$), which indicates that 
properly modeling the insufficiency and indiscrimination issues
are indeed necessary when working with distantly-labeled rationales. It is noteworthy that combining Order Loss and Gate Loss further improves the F1-score by as much as $2.8\%$, which is larger than any of their separate applications. This illustrates that the two new methods, aiming at different quality issues, can be applied together in a natural/integral form to jointly exploit distantly-labeled rationales.

Although our Order Loss method can bring noticeable improvement, $0.8\%$ in F1 than Base Loss, on SST, our Gate Loss and the Base Loss only achieves comparable performance with vanilla CNN. 
We believe the reason is that SST actually suffers from severe insufficiency issues. 
 There are only 0.85 rationale words per sentence in SST, but 4.66 rationales per sentence in ACE-2005. 
Given that $76.7\%$ sentences hold only 1 annotated rationale word, there is not much for the early-stop mechanism in our Gate Loss to do on SST. In this case, the Gate Loss boils down to the Base version. 
That is also why the Order Loss obtains significant improvement ($1.1\%$ more in F1 than Base Loss) on SST, which is designed to encourage those PINs to contribute to model training as well. 

\section{Analysis}

\subsection{Efficacy Analysis}
In order to understand the running mechanisms of our methods, we should look at what our methods have done with the non-perfect rationales.  
To this end, we examine the influence of our proposed losses by analyzing the average salience scores of two specific types of words in ACE-2005, \textit{event arguments} and \textit{gold triggers}, corresponding to the target of Order Loss and Gate Loss, respectively.

\textit{Arguments} refer to entities (mentions) involved in an event. They are not annotated as rationales by us, but previous studies show the importance of these words for event extraction \cite{event2018aaai}. We expect the Order Loss could maintain enough focus on them.

\textit{Gold triggers} in a sentence refer to the gold-standard event trigger annotations in ACE 2005, which are considered to indeed cause that sentence to be labeled as an event mention by the ACE annotators.  
 As the decisive factor for event detection, the gold triggers definitely serve as essential indicators, and consistently deserve high focus from the detection model. We will explore whether Gate Loss can successfully perceive them and render them lasting, sufficient focus.

\paragraph{Give Weight to Helpful Non-rationales}
\begin{figure}[t]
    \centering
    \includegraphics[width=0.44
    \textwidth]{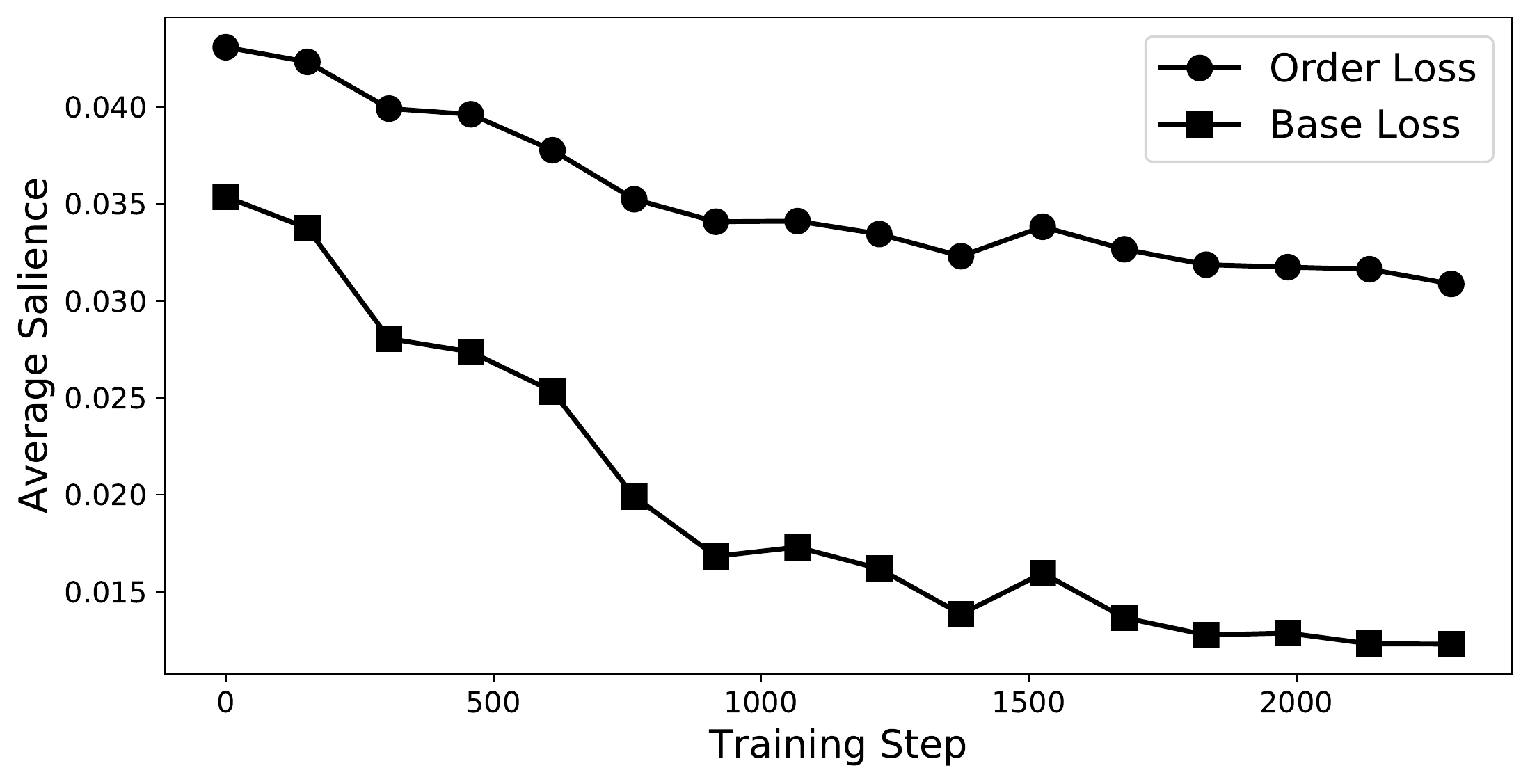}
    \caption{Average salience scores of argument words with Base Loss and Order Loss on ACE-2005. As training proceeds, Base Loss forces the argument to little focus, while salience scores in Order Loss maintain a high level. 
    }
    \label{fig:aver_gradient}
    \vspace{-0.25cm}
\end{figure}

We calculate the average salience scores of argument words on ACE-2005
with the Base Loss and Order Loss methods, respectively. 
As shown in Fig~\ref{fig:aver_gradient}, the average salience score of arguments when applying the Base Loss is much lower compared with the Order Loss during the whole training procedure. Equipped with the Order loss, the salience tends to stabilize at a high level. This illustrates that, unlike Base Loss, Order Loss allows arguments to obtain model emphasis. Thus, potential important words beyond rationales are able to contribute, making the model prediction more accurate.

\begin{figure}[t]
    \centering
    \includegraphics[width=0.44\textwidth]{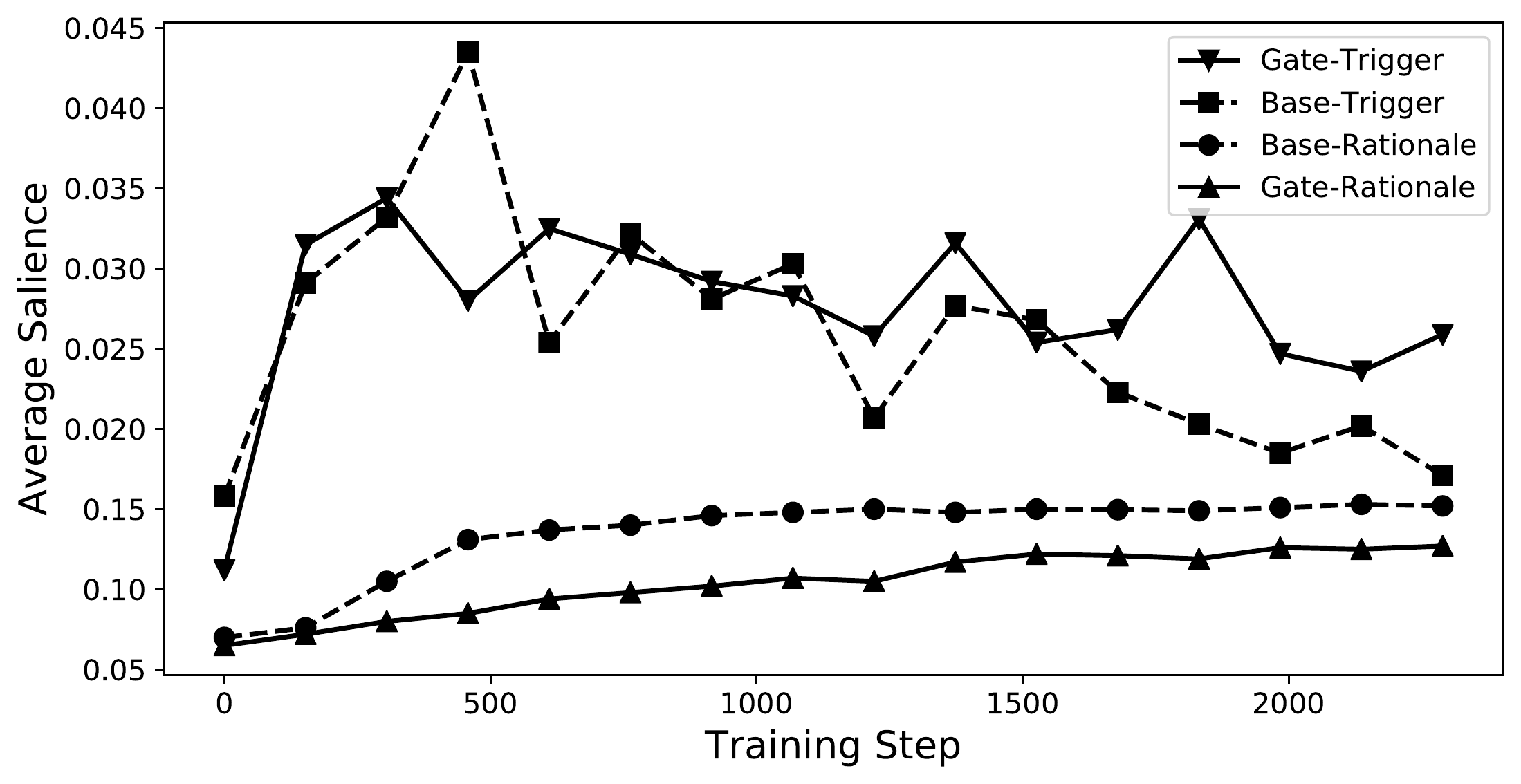}
    \caption{Average salience scores of crucial rationales (gold triggers) and the average salience of all rationales with Base Loss and Gate Loss on ACE-2005. After around 1500 training steps, Base Loss drives the salience of gold triggers towards average, while Gate Loss remains high discernment compared with Base Loss, with a higher focus on gold triggers and a lower average for all rationales.}
    \label{fig:prob_machanism}
\end{figure}

\paragraph{Focus on Crucial Rationales}
The average salience scores of gold triggers and all rationale words are plotted in Figure~\ref{fig:prob_machanism}. As can be seen, for both Gate Loss and Base Loss, the salience score of gold triggers increases quickly and surpasses the average salience scores of all rationales at the beginning.
However, the salience score of gold triggers in Base Loss begins to decline as training proceeds, to finally comparable with other rationales. In contrast, with Gate Loss, the salience of gold triggers remains rather stable at a high value. Such stability shows that the early-stop mechanism introduced by Gate Loss helps maintain the focus on these crucial rationales, instead of forcing them to approach average. 

\subsection{Robustness  Analysis}
\begin{figure}[t]
    \centering
    \includegraphics[width=0.42\textwidth]{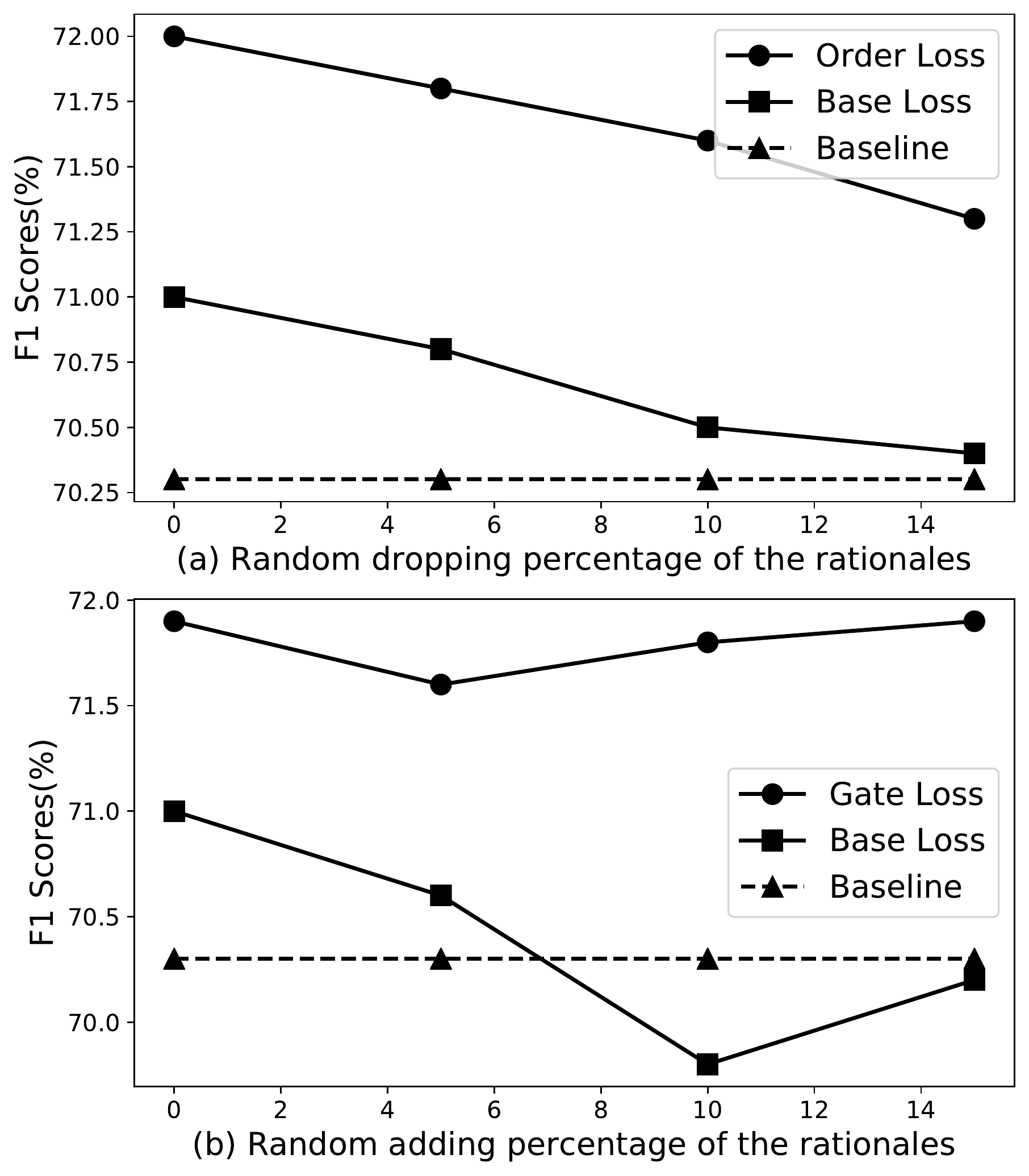}
    \caption{Results of perturbation experiments on ACE-2005. (a) shows the consequences of randomly removing 10\%, 20\%, 30\% of the distantly-labeled rationales, and (b) shows the effects of randomly adding 5\%, 10\%, 15\% extra words to rationale annotations.
    }
    \label{fig:control_exp}
    \vspace{-0.3cm}
\end{figure}

As shown in the previous section, our proposed methods can
alleviate the Insufficiency and Indiscrimination issues of the non-perfect rationales. Here, we take a step forward to 
the robustness of our proposed methods, e.g., how our methods will perform when given a much lower quality of rationales.

\paragraph{Working with scarcer rationales}

Now the question is: how our method will perform if the rationale-labeling rules are less inclusive and the rationales are even scarcer? 
To study the stability of the Order Loss, we create a more tough situation of scarcer rationales by gradually throwing away a small, random proportion of words from the original rationales on ACE-2005. 

Figure~\ref{fig:control_exp}(a) shows the performance of Order Loss under up to 30\% reduction of rationales, compared with Base Loss. We see that Order Loss undergoes only minor losses of performance as more rationale words are transferred to non-rationales, since Order Loss is designed to consider the PINs by spreading the model focus to those non-rationale words that can contribute to the final classification. However, Base Loss gradually loses its ability to incorporate priors, since it attempts to give the model focus entirely to the rationales, and finally slides to near baseline performance at around 30\% amount of perturbation. This indicates that our Order Loss can stably and efficiently learn from insufficient rationales while keeping an eye on other helpful words that are left out.

\paragraph{Working with noisy rationales}

A robust model should be capable of discerning whether a word is important indeed by itself, instead of simply checking the rationale label. We seek to examine whether the Gate Loss can still take effect under more and more severe pollution of false rationales. Specifically, we intentionally introduce noises to the rationale annotations by randomly labeling several non-rationale words in each case as ``\textit{rationale}'' in ACE-2005, and see how the performance of different methods is affected. 

As can be seen in Figure~\ref{fig:control_exp}(b), more noises do not pose a big threat to Gate Loss, with only a 0.6\% decline in its performance under at most 15\% amount of perturbation. However, Base Loss turns out to be highly dependent on the purity and reliability of the rationales, as its performance drastically falls to even below baseline within less than 10\% perturbation. This is not surprising, since Base Loss requires equally high attention on rationales, which is unreasonable for noisy rationale annotations. Nonetheless, the early-stop mechanism of Gate Loss circumvents overtraining on noises, thus outperforming the rigid requirements of Base Loss.

\section{Related Works}
Incorporating human priors has been well studied in different NLP applications with different forms of rationales. 
\citet{zaidan2007using} attains a more reliable Support Vector Machine by adding contrast training examples, which mask out important substrings. 
\citet{yu2019rethinking} exploit pre-annotated rationales to train an extractor and use the extracted words for classification. 
\citet{luo2018marrying} concatenates information of regular expressions to word embeddings for spoken language understanding. 
\citet{re2020emnlp} uses an RNN to model regular expressions for text classification tasks.
Most of these works provide effective ways to utilize word-level knowledge, but none of them formally considers the quality issues with the distantly-labeled rationales. Additionally, \citet{featurefeedback2017} discuss the insufficiency issue in the feature feedback framework, and try to incorporate vague feature feedback into a linear classifier. 

As a widely-used explanation method, the attention mechanism is often applied with constraints to guide model focus towards the significant part of inputs \cite{liu2017exploiting,nguyen2018killed,bao2018deriving}. Our proposed methods are currently based on gradient-based salience calculation, which is easier to obtain and model-agnostic, thus can be applied to a wider range with ease. But our methods do not depend on specific calculation methods for word salience, and can be easily transplanted to attention-based constraints, which we will leave for future work.

Recent studies have provided various techniques to constrain gradient-based word salience. \citet{ross2017right} forces the gradient of features, which are annotated non-helpful, to be zero, to alter the decision boundary of the model. \citet{liu2019incorporating} calculates $L_2$ distance between Path Integrated Gradients attribution for selected tokens and a target value in the objective function, to mitigate unintended bias in toxic comment classification and improve classifier performance in scarce settings. \citet{ghaeini2019saliency} requires the gradients of all rationales to be positive to encourage the model to focus on salient words.
The success of these works motivates us to further explore the impact of distantly-labeled rationales, which are easier to obtain but will bring challenges to previous methods as we have shown in experiments. 
In our method, we formally consider the insufficiency and indiscrimination issues, and design two losses to not only push the classification model to take care of those potentially important non-rationales, but also discriminatively focus on rationales to avoid over-training on those non-helpful annotations. 

There is another line of works that try to explicitly produce human-readable rationales during model learning. \citet{lei2016rationalizing} use reinforcement learning to identify keywords as rationales to improve model interpretability. \citet{eraser2020tacl} further constructs a benchmark dataset to engage the research about interpretable model design. While, our work is to examine how to better incorporate non-perfect rationales into neural network models, which is orthogonal to that line of research.

\section{Conclusions}
While distantly-labeled rationales are easy to obtain, they are often insufficient and indiscriminative, compared with high quality expert annotations. 
In this paper, we provide new perspectives on how to deal with such rationales, and propose two novel methods
to guide a classification model to learn from potentially important non-rationales while avoiding over-training on noisy annotations. 
Experiments on two NLP classification tasks show that our methods can effectively tackle the mentioned quality issues and are robust enough to exploit the non-perfect rationales even in more tough situations.
Our methods are not limited to specific salience calculations, we hope to explore more forms of word salience and rationales in the future.
We also expect our approaches to be beneficial in other scenarios where rationales are noisy and incomplete. This even includes scenarios when rationales are not distantly labeled, e.g., crowdsourced human annotations with low agreement \cite{sen-etal-2020-human}.

\section*{Acknowledgments}
We thank the anonymous reviewers for the helpful comments and suggestions. This work is supported in part by the National Hi-Tech R\&D Program of China (2018YFC0831900) and the NSFC Grants (No.61672057, 61672058).

 \bibliographystyle{acl_natbib}
\bibliography{anthology,acl2021}

\newpage

\clearpage

\section*{Appendix}
\subsection*{Different Implementation of Gate Loss}
Considering the sum of rationale words salience shows whether the rational words have gained adequate focus in total, we use Bernoulli distribution as a gate to control the constraint in our Gate Loss:
\begin{equation}
    L_{a\_gate}=Bern(1-\sum_{x_i \in S_R} s_i)\sum_{x_i \in S_R} (s_i-1)^2
    \label{app:ber}
\end{equation}

We have tried other ways to perform the gate. First, we try to use the sum of rationale salience as a weight directly. We define \textbf{Soft Gate Loss} as
\begin{equation}
    L_{a\_soft\_gate}=(\sum_{x_i \in S_R} s_i) \sum_{x_i \in S_R} (s_i-1)^2
    \label{app:soft}
\end{equation}

Another way is to use a threshold to determine whether the focus rational words gained are sufficient. we define \textbf{Marginal Gate Loss} as
\begin{equation}
    L_{a\_marginal\_gate}=\mathbf{I}(\sum_{x_i \in S_R} s_i \leq t) \sum_{x_i \in S_R} (s_i-1)^2
    \label{app:marginal}
\end{equation}
where $\mathbf{I}$ is an indicator function and $t$ is a predefined threshold.

The performance of Soft Gate Loss and Marginal Gate Loss is shown in Table~\ref{tab:gate_compare}. As can be seen, our Bernoulli gate performs best among all the three gate calculation methods. 
Soft Gate Loss can bring a bit more improvement than Base Loss, but not significant enough, which illustrates that a soft control may not be suitable. As for the Marginal Gate Loss, its performance is very sensitive to the selection of threshold and the best F1 is only $71.1\%$, which is still lower than Bernoulli Gate.

Thus, taking both performance and stability into consideration, we choose Bernoulli Gate as our implementation of Gate Loss.

\begin{table}[t]
\centering
\setlength{\tabcolsep}{16pt}
\begin{tabular}{ll}
\toprule
               & \multicolumn{1}{c}{F1} \\ \midrule
Baseline       & 0.698                  \\ \midrule
~~~+Base Loss     & 0.704                  \\ \midrule
~~~+Soft          & 0.707                  \\ \midrule
~~~+Marginal~(0.9) & 0.709                  \\
~~~+Marginal~(0.7) & 0.711                  \\
~~~+Marginal~(0.5) & 0.698                  \\
~~~+Marginal~(0.3) & 0.699                  \\ \midrule
~~~+Bernoulli     & 0.715                  \\ \bottomrule
\end{tabular}
\caption{The performance of different gate calculations on ACE-2005. +Soft, +Marginal, +Bernoulli means use Soft Gate Loss, Marginal Gate Loss and Gate Loss respectively. And the number in the bracket for Marginal Gate represents the threshold.}
\label{tab:gate_compare}
\end{table}

\end{document}